\pdfoutput=1

\documentclass[11pt]{article}

\usepackage{naacl2021}

\usepackage{times}
\usepackage{latexsym}
\usepackage{multirow}
\usepackage{graphicx}
\usepackage{xcolor}
\usepackage{float}

\usepackage[T1]{fontenc}

\usepackage[utf8]{inputenc}

\usepackage{microtype}

\title{Improving Data Driven Inverse Text Normalization\\ using Data Augmentation}

\author{Laxmi Pandey$^{\dagger}$, Debjyoti Paul, Pooja Chitkara, Yutong Pang, \\ \textbf{Xuedong Zhang, Kjell Schubert, Mark Chou, Shu Liu, Yatharth Saraf}\\
  University of California Merced$^{\dagger}$, Meta Inc. \\
}

\begin{document}
\maketitle
\begin{abstract}
Inverse text normalization (ITN) is used to convert the spoken form output of an automatic speech recognition (ASR) system to a written form. Traditional handcrafted ITN rules can be complex to transcribe and maintain. Meanwhile neural modeling approaches require quality large-scale spoken-written pair examples in the same or similar domain as the ASR system (in-domain data), to train. Both these approaches require costly and complex annotations. In this paper, we present a data augmentation technique that effectively generates rich spoken-written numeric pairs from out-of-domain textual data with minimal human annotation. We empirically demonstrate that ITN model trained using our data augmentation technique consistently outperform ITN model trained using only in-domain data across all numeric surfaces like cardinal, currency, and fraction, by an overall accuracy of 14.44\%.
\end{abstract}

\section{Introduction}
Inverse Text Normalization (ITN) is used to convert spoken form output from an automatic speech recognition (ASR) system to the corresponding written form.
ITN can be challenging since multiple different spoken forms can express identical written expressions. For example, both twenty twenty (the year) and two thousand twenty (numeric) can be transcribed to `2020'. Conversely, the same spoken form can be transcribed to two or more different written expressions depending on the context. For example, twenty twenty can be transcribed to 2020 (for the year), to 20/20 (to denote eye vision), or to 20:20 (to represent time). Such a many-to-many mapping between spoken and written forms and dependence on context makes ITN an interesting and challenging problem in speech recognition. In Table \ref{table2} of the Appendix, we present additional examples of spoken-written pairs using ITN.

\begin{table}[t]
{\small
\centering
\begin{tabular}{l|l}
\hline
\textbf{Spoken form}              & \textbf{Written form}    \\ \hline
do you like nineties music        & do you like 90s music    \\ \hline
let's meet at three thirty  & let's meet at 3:30 \\ \hline
three thirty kilos  & 330 kilos \\ \hline
he is at thirty percent of his goal & he is at 30\% of his goal \\\hline
\end{tabular}
\vspace{-2mm}
\caption{Examples of Spoken-Written pairs.}
\label{table1}
\vspace{-7mm}
}
\end{table}

Traditional ITN systems rely on hand-curated rules per language which are then translated into weighted finite-state transducer (FST) grammars \cite{ebden2015kestrel} to perform inverse normalizations. However, handwritten rules can be complex to transcribe and maintain, and context can be difficult to encode in FSTs.

There has been a renewed interest in the deep learning community to explore data-driven approaches to ITN. A popular ITN approach is to use a set of simple hand-written rules together with a neural model that can statistically learn how and when to apply these rules \cite{ihori2020large, mansfield2019neural, pramanik2019text, shugrina2010formatting, sak2013written, ju2008language}. The rules needed in a neural network model-based system are more straightforward to produce than a purely rule-based FST-compiled system. However, sufficient training data in the same or similar domain as the ASR (in-domain data) is needed to train a neural model. 

To generate training data for ITN models, a common approach is to use a text normalization (TN) system \cite{zhang2019neural}. However, since the TN system only outputs one flawless spoken form per written input, it does not cover the variations of spoken forms that can be generated from a single written form. If we train a model with the over simplified spoken-written pairs, the model usually over-fits to the TN system, reflecting high accuracy for the curated entities, but the model can struggle to generalize to real-world use cases.

To this end, we make the following contributions in the paper: \\ \vspace{-7mm}
\begin{itemize}
    \item We propose a method of robust numeric data augmentation that can process out-of-domain text-only data to generate spoken-written pairs with large variations of spoken outputs per written form, to train a data-driven ITN model. \\ \vspace{-8mm}
    \item We train and evaluate a multitask sequence-encoder ITN model to empirically measure the effectiveness of our approach. Experiments demonstrate a significant improvement in accuracy across all numeric labels such as cardinal, currency, and fraction, with our data augmentation approach. \\ \vspace{-6mm}
\end{itemize}

\section{Data Driven ITN}
\vspace{-2mm}
Our proposed data-driven ITN system comprises of two components: {\em (a)} A \textit{data augmentation} module that takes a written form sequence as input and outputs their corresponding spoken forms. {\em (b)} A multitask sequence encoder-based \textit{ITN model} module that is first pretrained on data-augmentation dataset, and then finetuned with a limited amount of in-domain human-revised spoken-written pairs dataset.

\subsection{Data Augmentation}\label{AA}
Considering that people can speak a particular written form in multiple ways in the real world- for example,  {\em 5.0} could be verbalized as {\em five point zero}, {\em five point o}, {\em five dot zero}, etc. - we have developed a specialized data augmentation method for data-driven ITN modeling. 

\begin{figure}[t]
{\small
\centering
\includegraphics[width=0.5\textwidth]{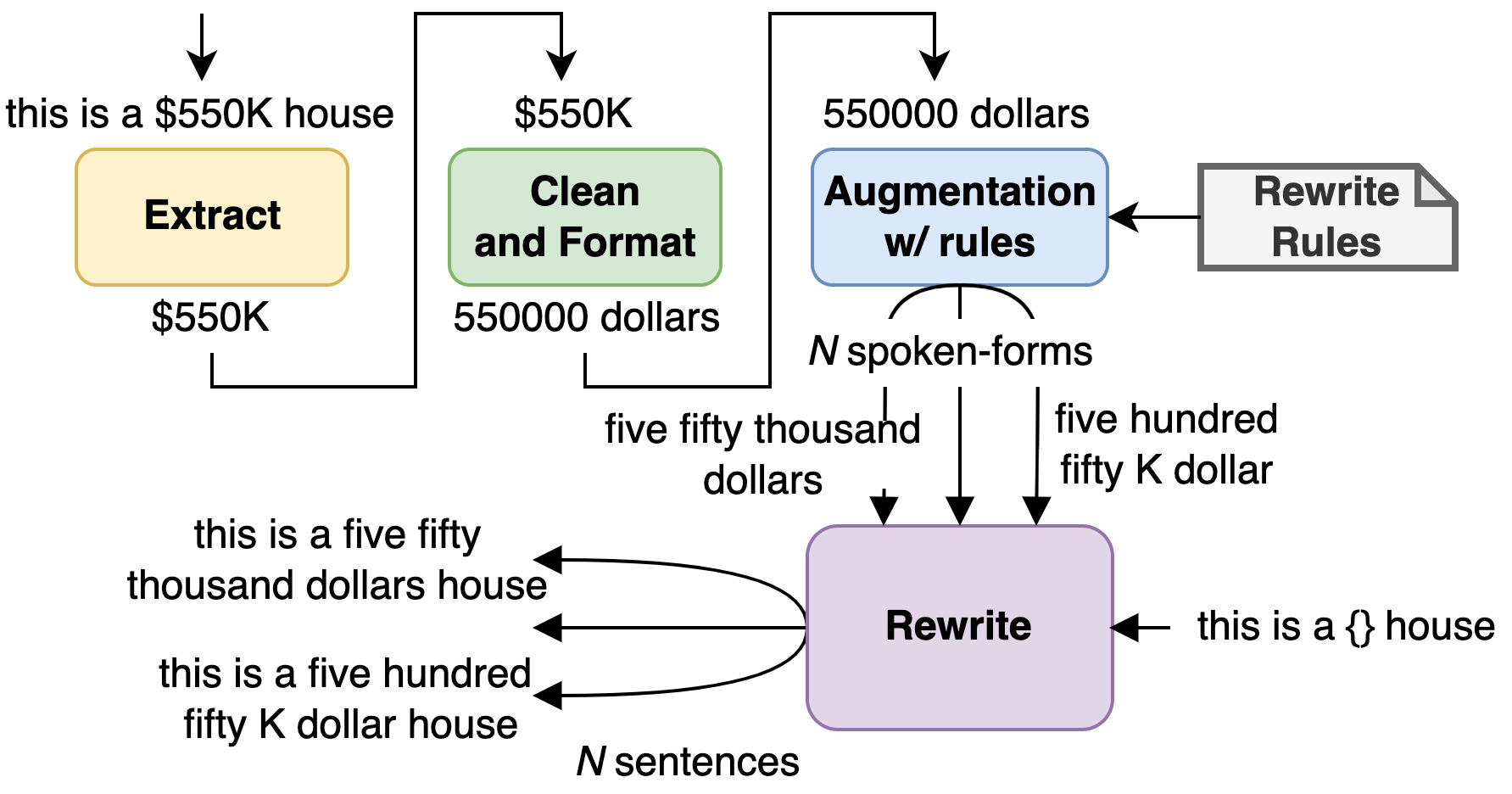}
\vspace{-4mm}
\caption{Specialized data augmentation system for our ITN modeling.}
\label{fig:augmentation}
}
\vspace{-6mm}
\end{figure}

Unlike a conventional written to spoken TN system, our ITN augmentation system is capable of generating diversified spoken forms by introducing almost all possible spoken variations to the written forms as shown in Table \ref{table2}. Our augmentation system performs a series of steps for each written text as input; depicted in Figure \ref{fig:augmentation} as follows.

\vspace{-3mm}
\begin{itemize}
    \item[a.] Extract text chunks matching cardinal, ordinal, currency, fraction, measures, abbreviations, phone numbers, and time entities, etc. \\ \vspace{-8mm}
    \item[b.] Entities matching text chunks are then cleaned and formatted. E.g., {\em Time: 12:45 $\rightarrow$ 12 hours 45 minutes.}, {\em Measures: 10K lb $\rightarrow$ 10000 lb.}  \\ \vspace{-8mm}
    \item[c.] The data augmentation core generates multiple spoken forms per written form of formatted text input with the help of rewrite rules. The rewrite rules are specialized pattern mappings from written to spoken forms; augmentation core applies these rules on texts (recursively, if required) with exhaustive combinations. \\ \vspace{-8mm}
    \item[d.] Finally, the rewrite module replaces the written-form text in the original sentence with $N$ generated diverse spoken forms. \\ \vspace{-6mm}
\end{itemize}

To measure the diversity or assortment of our data augmentation system in 
comparison to the conventional TN system, we use the following equation 
$f_{diverse}$.

\vspace{-2mm}

\begin{equation}
    f_{diverse} = \frac{ |\textit{spoken form entities}|}{|\textit{written form entities}|}
\end{equation}

We found that the augmentation system on our social networking comments dataset generates 22.64$\times$ more diverse data than the baseline. We present a few example utterances from the data augmentation system in Table \ref{table2}.


\begin{table*}[t]
\vspace{-9mm}
{\small
\centering
\begin{tabular}{c|c|c|c|c|c|c|c}
\hline
\multirow{2}{*}{\textbf{\begin{tabular}[c]{@{}c@{}}Spoken \\ Input\end{tabular}}} & \multicolumn{5}{c|}{\textbf{Label}}                                                          & \multirow{2}{*}{\textbf{Post-processing}} & \multirow{2}{*}{\textbf{\begin{tabular}[c]{@{}c@{}}Written \\ Output\end{tabular}}} \\ \cline{2-6}
                                                                                  & \textbf{Rewrite} & \textbf{Prepend} & \textbf{Space} & \textbf{PostStart} & \textbf{PostEnd} &                                           &                                  \\ \hline
i                                                                                 & None             & None             & On             & None               & None             & i                                         & i                                \\ \hline
have                                                                              & None             & None             & On             & None               & None             & have                                      & have                             \\ \hline
one                                                                               & Cardinal         & None             & On             & MajorCurrency      & None             & \textless{}MajorCurrency\textgreater 1    & \multirow{3}{*}{\$120}           \\ \cline{1-7}
twenty                                                                            & Cardinal         & None             & Off            & None               & None             & 20                                        &                                  \\ \cline{1-7}
dollar                                                                            & CurrencySymbol   & None             & Off            & None               & MajorCurrency    & \$\textless{}MajorCurrency\textgreater{}  &                          \\ \hline
\end{tabular}
\vspace{-2mm}
\caption{Running example of converting spoken input to labels for training followed by written spoken form to its corresponding written form.}
\label{table3}
}
\vspace{-5mm}
\end{table*}

\subsection{ITN Modeling}\label{BB}
\begin{figure}[t]
\centering
\includegraphics[width=\linewidth]{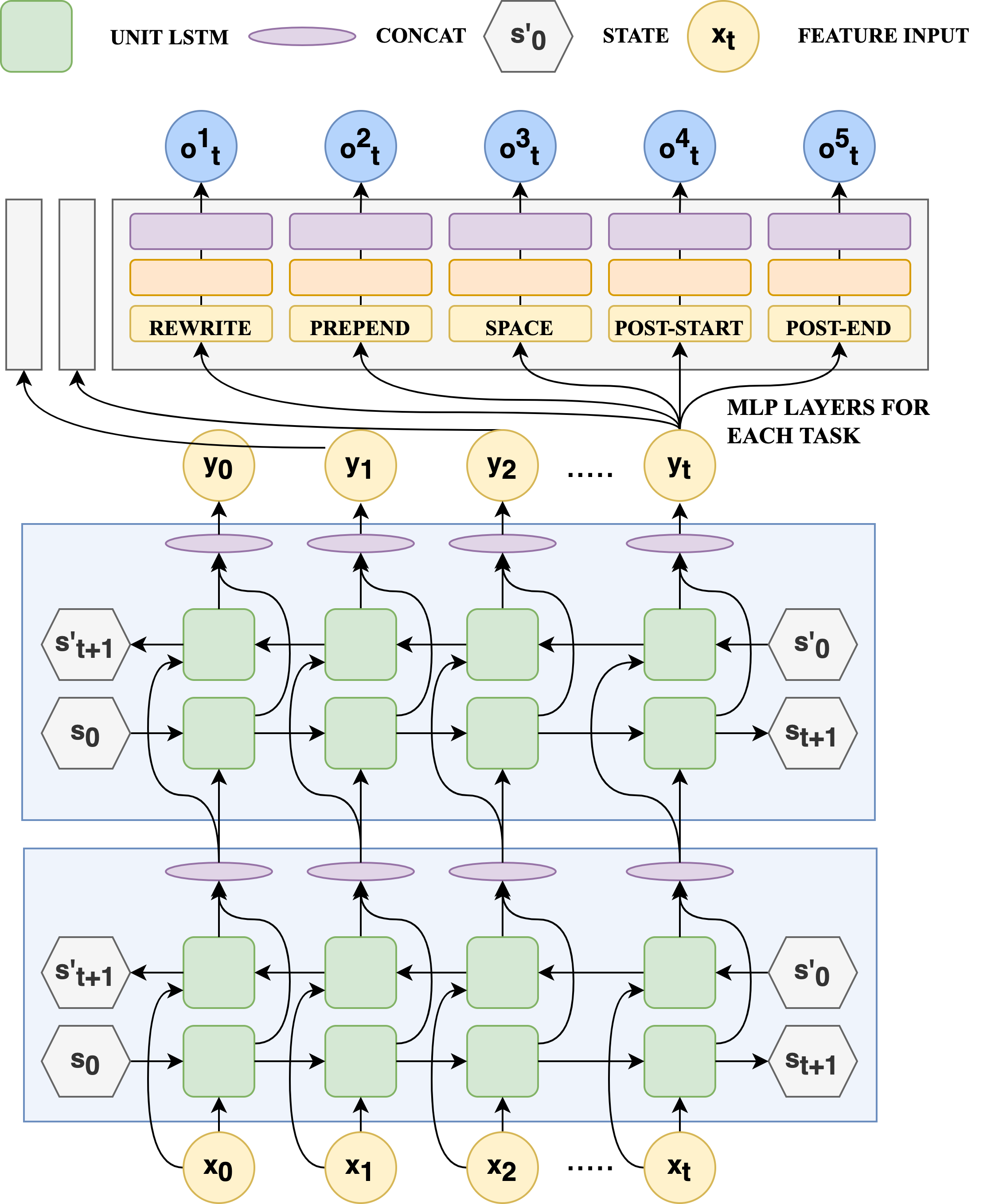}
\caption{ A multitask multilabel bidirectional LSTM sequence encoder with five tasks for ITN.}
\label{model}
\end{figure}

With the data augmentation system in place, we can generate a significant number of spoken-written numeric pairs in our dataset. Unlike the Sequence-to-Sequence Encoder-Decoder \cite{sutskever2014sequence} modeling approach, we use a Sequence Encoder-based multitask classification model inspired from \cite{pusateri17_interspeech}. It has a couple of advantages over the former model, namely (a) Encoder-only architecture reduces complexity and computation of the model and makes it more amenable to deploy with on-device ASR models where memory restrictions are critical, (b) End to end sequence-to-sequence models can be hard to debug and less interpretable. With the encoder label classifier approach, model architects have finer control over hot fixes over ITN entities.
Our ITN model, pictorially represented in Figure \ref{model}, takes a sequence of sentence pieces, and inputs a feature vector per piece, represented as $x_t$. Two layers of Bidirectional LSTMs take these features, encode them as $y_t$, and then use a multi-layer perceptron stack for each classification task to output labels $o^{1-5}$. The five classification tasks for ITN are: {\em (a) Rewrite, (b) Prepend, (c) Space, (d) Post-Start, (e) Post-End}, described in Sec \ref{sec:appendix}. We have a running example of label inference from spoken input form with these five tasks in Table \ref{table3}.

We use a label inference engine \cite{pusateri17_interspeech} with space tokenization replaced by sentence piece-based tokenization to facilitate generating training datasets with labels. The tokenization adheres to the vocabulary of rewrite rules the label inference uses, and keeps ITN specific pieces entirely, e.g., {\em ninety} is not tokenized as [{\em \_nine,ty}]. 
We intuitively expect languages like Hindi, German, Italian benefit from this strategy as higher cardinals in those languages are not separated by spaces, e.g., eighty seven $\rightarrow$ siebenundachtzig in German.

\begin{table*}[h]
\centering
\begin{tabular}{c|c|c|c}
\hline
\textbf{Entity} & 
\textbf{Baseline} &  \textbf{\begin{tabular}[c]{@{}c@{}}With data augmentation \\  (Change \%)\end{tabular}}  \\ \hline
Overall                                                                        & 27.15                                                                                       & 31.08 (14.44\%)                                                                                                             \\ \hline
Cardinal                                                               & 31.14                                                                                & 37.37 (19.99\%)                                                             \\ \hline
Currency                                                                 & 40.08                                                                 & 42.83 (6.85\%)                                                                                \\ \hline
Fraction                                                                & 5.96                                                                               & 16.14 (170.59\%)                                                                              \\ \hline

\end{tabular}
\caption{Accuracy performance comparison of ITN candidate model with baseline (in \%).}
\label{table4}
\vspace{-4mm}
\end{table*}

\subsection{Datasets}\label{DD}
\vspace{-1mm}
\paragraph{Source domain data.}
We use an aggregated and de-identified English social media text corpus as source domain data for ITN model training, containing a random sample of 110 million posts and comments.
\vspace{-2mm}
\paragraph{Target domain data.}
It is comprised of a small human-supervised (annotated) crowdsourced dataset of around 50K sentences from dictation and assistant domains. We use a part of this dataset as an evaluation dataset, generated with multiple-pass human reviews that ensure the highest quality.
\vspace{-2mm}
\paragraph{General domain data.}
We use this dataset to train the embedding layer in the ITN model. This dataset is from Fischer English training speech transcripts \cite{fischer2004}. 

\begin{figure}[t]
\vspace{-2mm}
\centering
{\small
\includegraphics[width=0.5\textwidth]{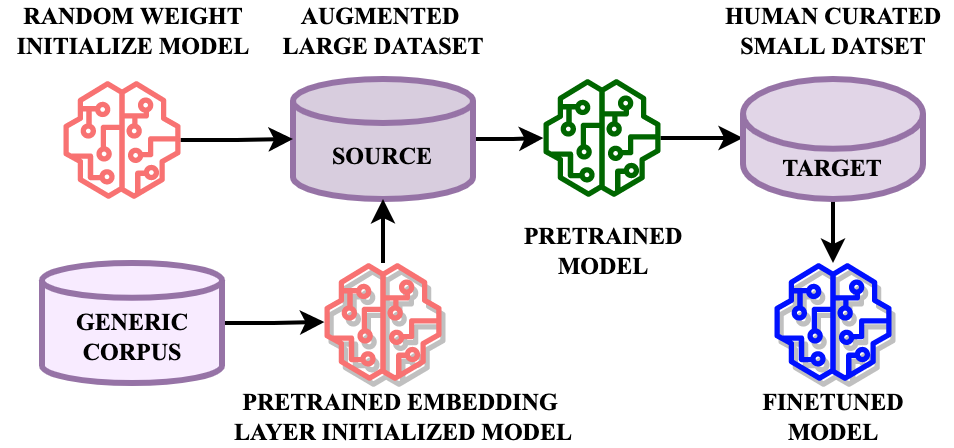}
\vspace{-1mm}
\caption{An overview of the proposed domain adaptation approach with source pretraining and general embedding pretraining.}
}
\vspace{-6mm}
\end{figure}

\vspace{-2mm}
\subsection{ITN training}\label{CC}
Before training the ITN model, we attempt to quantify the dissimilarity of the target domain data (human supervised) and source domain data (social media comments). To achieve this, we extract the top N($\approx$10K) most frequent unigram and bigram samples (excluding stop-words) from each dataset, and compare the vocabulary overlap across these N-gram samples. The smaller the overlap, the larger the difference in the domains will be, and the higher the potential for domain adaptation is. Results shown in Figure \ref{fig:overlap} reveal that each domain is substantially different from the other. The spoken forms of the target and source domains match only 26.8\% and 14.7\% for unigrams and bigrams respectively. Similarly, the written forms of the target and source domains match 31.8\% and 23\% for unigram and bigrams respectively.

We propose a strategy of pre-training with augmented source domain data followed by fine-tuning with target domain data. First, we train a language model using general domain data to initialize the embedding layer of the ITN model. We use the augmented spoken-written source domain dataset to train the ITN model. 
Then we finetune this model with the target domain data until the model converges.

\section{Experiments}
\vspace{-1mm}
\subsection{Scoring Metrics}
\vspace{-1mm}
We used accuracy as a metric to benchmark our ITN models. For each ITN type, 
we follow the following steps: \\ \vspace{-7mm}
\begin{itemize}
    \item First, align token pairs from written reference ({\em ref}) from evaluation dataset and model hypothesis ({\em hyp}) pairs.\\ \vspace{-8mm}
    \item For all {\em ref-hyp} pairs, we extract the entity types. We consider cardinal, currency, and fraction types since they are easy to match with regular expression patterns. In the future, the list of entity types can be extended for evaluation.  \\ \vspace{-8mm}
    \item Accuracy is calculated based on {\em hyp} and {\em ref} correctness.\\ \vspace{-6mm}
    \begin{equation}
    \textit{accuracy}= \frac{|\textit{correct}|}{|\textit{correct+ error}|}    
    \end{equation}
\end{itemize}

\subsection{Results}
\label{II}
    We explore ITN training with augmented data (section \ref{CC}) and demonstrate its efficacy on the target human supervised datasets. We also restrict our model size to less than 2MB with feature embedding dimensions of 64 and LSTM hidden size of 256. Table \ref{table4} presents the performance comparison of each investigated technique. Below are the descriptions of experiments:\\
       \hspace{0.5cm}$\bullet$ \textbf{Baseline:} Model trained on limited amount of human supervised dataset without augmentation.\\
       \hspace{0.5cm}$\bullet$ \textbf{Candidates:} Model is pre-trained with a large amount of augmented data and finetuned with a limited amount of human supervised data in target domain with pre-trained embedding layer initialization \cite{kocmi2017exploration}. 
       
   We use source domain data with 80:20 split as a training-validation set. Then we evaluate ITN models on a high-quality human supervised dataset. Table \ref{table4} presents the comparative performance of the proposed ITN model against the baseline. We observe a significant improvement in the proposed data-driven ITN model compared to the baseline with a 14.44\% improvement in accuracy. We are able to improve the performance on all entity types - cardinal, currency, and fraction.

\vspace{-2mm}
\section{Conclusion}
\vspace{-2mm}
In this paper, we introduce a robust data augmentation methodology for ITN that can generate rich and variant spoken-written pairs from out-of-domain textual (written) data for numeric entity types. We empirically demonstrate that our technique significantly improves ITN in its target domain. We believe this methodology can be particularly helpful for ITN in areas where training data in the same domain as the upstream ASR system is not readily available, and we hope it will encourage greater exploration of data-driven ITN methodologies in such areas of spoken technologies.
\bibliography{custom}
\bibliographystyle{acl_natbib}

\appendix

\section{Appendix}
\label{sec:appendix}
{\small

{\small Classification Task Labels.} We define the five classification tasks for the ITN model here.
\vspace{-2mm}
\begin{itemize}
\item {\bf Rewrite} indicates if a substring/string/word-piece needs to be rewritten with another. Rewrite labels are classified based on their types and actions: {\em cardinal, cardinal-decade, cardinal-hundred, cardinal-thousand, cardinal-million, cardinal-billion, ordinal, verbatim, abbreviate, measure, currency} etc. \\ \vspace{-6mm}
\item {\bf Prepend} task adds strings, digits or symbols in the beginning of a substring/word-piece. Example labels are {\em period, colon, slash, hyphen, digits} etc. \\ \vspace{-6mm}
\item {\bf Space} controls the addition of white-spaces before the reference word-piece. \\ \vspace{-6mm}
\item {\bf Post Start} and {\bf Post End} marks the beginning and end of a string chunk for rewrite post-processing. Types of post-processing labels are {\em currency, measure, magnitude} etc.\\ \vspace{-6mm}
\end{itemize}
}
\begin{table}[H]
{\scriptsize
\centering
\begin{tabular}{l|l|l}
\hline
\textbf{Input} & \textbf{Conventional TN}  & \textbf{Our Augmentation System} \\ \hline
123            & \begin{tabular}[c]{@{}l@{}}one hundred \\ twenty three\end{tabular}    & \begin{tabular}[c]{@{}l@{}}one hundred twenty three\\ one twenty three\\ one hundred and twenty three\\ one two three\end{tabular}                                                                                                                                                          \\ \hline
\$123          & \begin{tabular}[c]{@{}l@{}}one hundred  \\ twenty three dollars\end{tabular} & \begin{tabular}[c]{@{}l@{}}one hundred twenty three dollars\\ one hundred twenty three dollar\\ one twenty three dollars\\ one twenty three dollar\\ one hundred and twenty three dollars\\ one hundred and twenty three dollar\\ one two three dollars\\ one two three dollar\end{tabular} \\ \hline
123g           & \begin{tabular}[c]{@{}l@{}}one hundred \\  twenty three grams\end{tabular}    & \begin{tabular}[c]{@{}l@{}}one hundred twenty three grams\\ one hundred twenty three gram\\ one twenty three grams\\ one twenty three gram\\ one hundred and twenty three grams\\ one hundred and twenty three gram\\ one two three grams\\ one two three gram\end{tabular}                 \\\hline
\end{tabular}
\caption{Examples of generated spoken form using conventional TN system and the developed augmentation system}
\label{table2}
}
\vspace{-6mm}
\end{table}

\begin{figure}[H]
{\small
\centering
\includegraphics[width=0.5\textwidth]{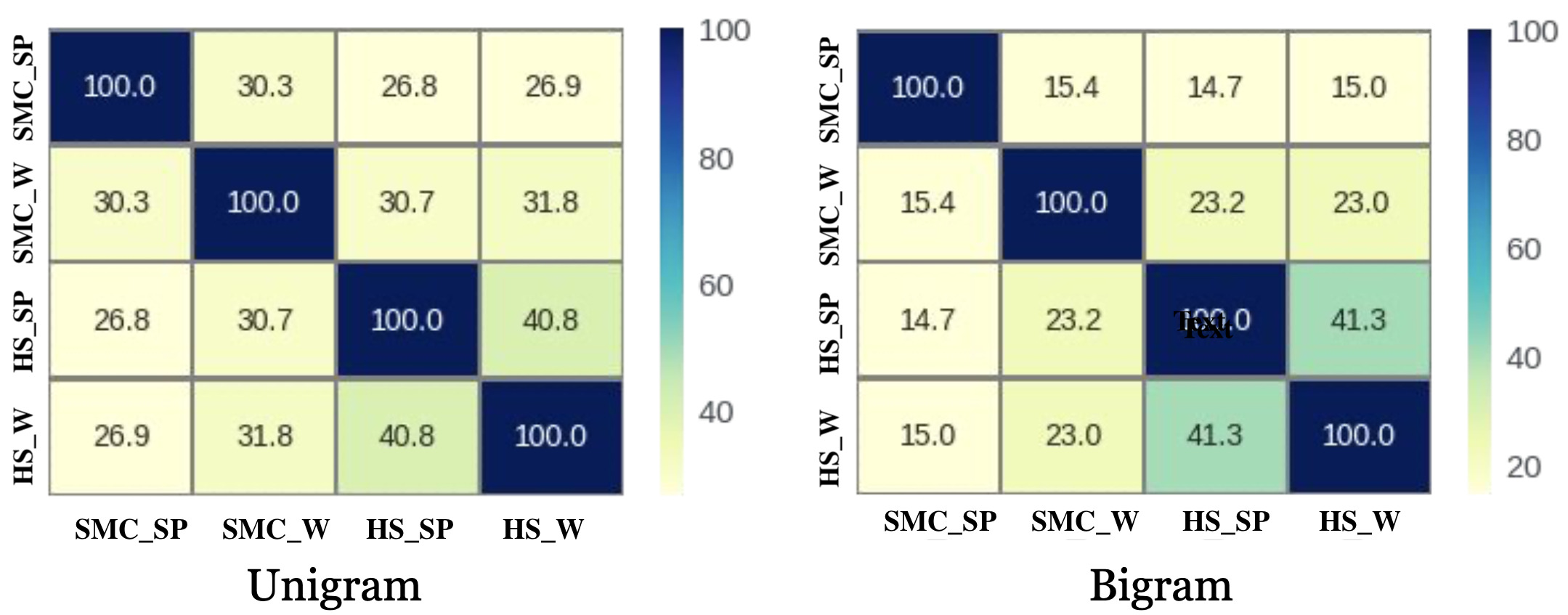}
\vspace{-6mm}
\caption{The figure shows Unigram and Bigram vocabulary overlap (\%) between {\em source} social media comments (SMC) and  {\em target} human supervised (HS) datasets for both spoken (SP) and written (W) based tokens. This represents how the distribution of tokens varies for different domains and shows the requirement of domain adaptation.}
\label{fig:overlap}
}
\end{figure}

\end{document}